\newcounter{myctr}

\documentclass{ws-acs}
\usepackage{url}
\usepackage{epsfig}
\usepackage{algorithm}
\usepackage[noend]{algorithmic}

\usepackage{algorithm}
\usepackage[noend]{algorithmic}
\usepackage{graphicx}
\usepackage{subfigure}
\usepackage{amssymb}
\usepackage{amsmath}
\usepackage{stmaryrd}

\newcommand{\G}{\mathcal{G}}
\newcommand{\J}{\mathcal{J}}
\newcommand{\T}{\mathcal{T}}

\newcommand{\I}{\mathcal{I}}
\newcommand{\SG}{\mathcal{S}_{\G}}
\newcommand{\ST}{\mathcal{S}_{\T}}

\newcommand{\AD}{App}
\newcommand{\DD}{Dis}

\newcommand{\TVG}{\ensuremath{\G=(V,E,\T,\rho,\zeta)}}

\begin{document}

\makeatletter
\def\@biblabel#1{[#1]}
\makeatother

\markboth{Francesca Giardini, Walter Quattrociocchi, Rosaria Conte}{Understanding opinions. A cognitive and formal account}

%
\catchline{}{}{}{}{}
%

\title{Understanding opinions. A cognitive and formal account}


\author{Francesca Giardini}
\address{Department of Cognitive Science, Central European University,\\
Budapest, Hungary, email: GiardiniF@ceu.hu
\footnote{Central European University, Hungary email: GiardiniF@ceu.hu}\\
GiardiniF@ceu.hu}

\author{Walter Quattrociocchi}
\address{Department of Mathematics and Computer Sciences,\\
University Of Siena, Italy
\footnote{University of Siena, Italy email: walter.quattriocchi@unisi.it}\\
walter.quattriocchi@unisi.it}

\author{Rosaria Conte}

\address{LABSS,\\
CNR - Institute of Cognitive Sciences and Technologies, Rome, Italy
\footnote{Labss-ISTC-CNR, Italy email: rosaria.conte@istc.cnr.it}\\
rosaria.conte@istc.cnr.it}

\maketitle

\begin{history}
\received{(received date)}
\revised{(revised date)}
\end{history}

\begin{abstract}
The study of opinions, their formation and change, is one of the defining topics addressed by social psychology, but in recent years other disciplines, as computer science and complexity, have addressed this challenge. 
Despite the flourishing of different models and theories in both fields, several key questions still remain unanswered. The aim of this paper is to challenge the current theories on opinion by putting forward a cognitively grounded model where opinions are described as specific mental representations whose main properties are put forward. A comparison with reputation will be also presented.
\end{abstract}


\section{Introduction}
\label{sec:1}
Opinions represent a conspicuous part of our mental representations. A large part of our social time is spent in exchanging, evaluating, revising and comparing our opinions. We also say, about many different issues, that we have opinions and we try to convince others about the groundedness of our own opinions. Since the beginning of the last century, social psychologists have been interested in understanding the specificity of opinions, as compared to other kinds of mental representations, by focusing their attention on the multiplicity of dimensions, including attitudes, beliefs and evaluations, that take part within this phenomenon. 
Also political science has always been very attentive to what is considered as a way to measure people's preferences and beliefs about publicly relevant issues. Many of these contributions have been directed towards understanding the so-called \emph{public opinion} and the processes through which it is possible to influence it, manipulating people's awareness and tendencies (\cite{lippmann}). 
More recently, other disciplines have shown a great interest regarding such an issue, ranging from computer science passing through socio-physics (\cite{Castellano2007,Galam08}) up to complexity science (\cite{Lorenz07}). 

Despite the large amount of studies on opinions, the term itself and the underlying concept are poorly specified and too general, since there are at least two classes of mental representations that can be termed “opinions” but they differ with regard to important aspects. Moreover, relevant contributions coming from social psychology and computer science try to model distinct issues, thus making the analysis of opinions quite difficult. This lack of sound theoretical contributions is often compensated by giving more preeminence to transmission and communication processes, thus partially putting aside the “ontological” issue.
In this work we propose a theoretical account in which, starting from a critical review of approaches coming from social psychology and computer science, the necessity of a cognitive approach is claimed. Defining the specific cognitive features that characterize an opinion, thus distinguishing it from other mental representations, and introducing also two different kinds of opinions, evaluative and factual, we will claim for the necessity of investigating the mental roots of opinions, in order to understand how they are transformed and manipulated within and between minds. This means that an opinion is specific with regard to other mental representations, that has special features and is transformed through specific mental processes. Defining an opinion in terms of its mental ingredients permits to predict opinion change, its persistence, the effects of contrasting forces and alternative paths of diffusion, because the different forces are endogenously determined by specific rules. Understanding opinions, describing how they are generated and revised, and how fare opinions travel over the social space both as a consequence of social influence and as one of the main means through which social influence unfolds, is crucial for grasping a deeper understanding of human social cognition and behaviors. 
Moreover, our cognitive analysis is supported by a preliminary formal description, in which a new tool called {\em Time Varying Graphs} \cite{CFQS2010} is presented. This formalism has been developed to deal with dynamically evolving systems\cite{AQ2010a,AQ2010b}, and it allows to overcome some of the limitations imposed by other instruments -e.g. metrics, formalisms – that are not suited to account for a) the relationships between opinions and other epistemic representations and b) their dynamics both at social and individual level.
In section 2, a critical introduction to some of the main contributions about opinions is provided. Section 3 is devoted to the description of our model, in which a definition of opinions as specific mental representations and cognitively founded hypotheses about their diffusion and change will be put forward. In section 4 a preliminary formal account of how opinions are generated and how they can change is provided. In section 5 some conclusions are drawn and future directions are suggested. 

\section{A critique to existing approaches}
The understanding of opinions requires to take into account two levels of explanation: the individual and the social level. 
As mental representations, opinions are created within agents' minds and they need to be integrated with the existing network of beliefs, data, information, memories and evaluations. However, in opinion change the social influence plays a major role and the sharedness of an opinion can heavily affect its persistence and resistance to change. These two dimensions are tightly linked and their interplay is one of the defining features of opinions, but social psychology and computer science are usually interested in tackling only one of these two aspects, without studying them in combination. We claim that developing a cognitive theory of opinions allows us to combine the micro- and the macro-level, understanding how macro-social phenomena emerge, unintentionally, from micro-elements and their interactions. In this way we can see that opinions derive from agents' cognitive representations and states but they also exist in the social space, in which they are transmitted and shared, and this social process affects, in turn, individuals' opinions. This complex loop requires a non-reductionist approach in order to deal with both levels, without giving preeminence to one or the other. 
 
Social psychology mainly focuses on the individual side, trying to describe how opinions are generated within the mind, devoting much attention to define attitudes and evaluations, but paying little attention to the socially interactive dimension of opinions. On the other hand, scholars from computer science and physics have tried to explain how different opinions can coexist or how they are modified through communication, treating opinions as mere objects that are exchanged and revised according to certain mechanisms that are quite far from the reality of cognitive and social processes. 
In both cases there is a reductionist fallacy that works in different ways but in both cases results in a downgrading of a complex issue into either a set of unrelated specific elements or a unidimensional object that is far from the complexity of a cognitive representation. 

\subsection{Social psychology: individualistic fallacy}
Social psychologists have devoted much attention to the study of opinions' formation and spreading, but a comprehensive and definite model allowing for an operational and generative account is still missing. Providing a comprehensive review of social psychology literature is beyond the scope of this work, but in this section we will discuss some of the main theories in order to underline how partial is the picture of opinions emerging from these studies. 

In general, opinions are treated as synonyms for different mental objects, as beliefs \cite{fishbeinAjzen}, or more frequently, attitudes. Opinions are often conceptualized as attitudes \cite{mcguire}, \cite{olsonZanna}, \cite{Price} or they are used as interchangeable terms that have in common the fact of being affected by social influence and persuasion \cite{Pettyetal}. It is worth noticing that many contributions are specifically oriented to understand "public opinion" \cite{lavineLatane}, as the integration of opinions and attitudes coming from different sources and exposed to different kinds of influencing. Another general feature of the social psychology approach to opinions is the preeminence given to measuring opinions, rather than on conceptualizing them. As a result, many studies (for a review, see Schwarz N, Sudman S, eds. 1996. Answering Questions: Methodology for Determining Cognitive and Communicative Procesess in Survey Research. San Francisco: Jossey- Bass) tried to develop reliable and fine-tuned ways to measure people's approaches to general questions, partially abandoning the issue of defining what an opinion is and focusing on how it should be measured. 

Allport \cite{allport} recognizes the difference between attitudes and opinions but he nonetheless considers the measurement of opinions as one way of identifying the strength and value of personal attitudes. An alternative view contrasts the affective content of attitudes with the more cognitive quality of opinions that involve some kind of conscious judgments \cite{Fleming}. In general, it is possible to identify two main trends in the relevant literature: one more focused on attitudes and the other more centered on conscious reasoning and judgment.
Crespi \cite{crespi} considers individual opinions as " judgmental outcomes of an individual's transactions with the surrounding world" (p.19), emphasizing the interplay between what he calls an attitudinal system and the external world characterized by the presence of other agents and different subjective perceptions. Opinions are the outcomes of a judging process but this does not mean that they are necessarily rational or reasoned, although Crespi recognizes that they need to be consistent with the individual's beliefs, values and affective states.

As other authors already pointed out \cite{Masonetal.}, many models of opinion and social influence do not provide careful definitions of what an opinion is and how it is affected by social influence. This happens to be true also for theories of persuasion, like the social impact theory \cite{latane81}, a static theory of how social processes operate at the level of the individual at a given point in time. Part of this theory has been developed using computational modeling by Nowak, Szamrej and Latané \cite{Latane90}. In their model, individuals change their attitudes as a consequence of other individuals' influence. In parallel with the idea that social influence is proportional to a multiplicative function of the strength, immediacy, and number of sources in a social force field \cite{latane81}, \cite{lavineLatane} suggest that each attitude within a cognitive structure is jointly determined by the strength, immediacy, and number of linked attitudes as individuals seek harmony, balance, or consistency among them. Although very interesting, this account fails to distinguish between attitudes and beliefs and does not explain how inconsistencies can be resolved.
The effect of communication on opinion formation has been addressed by different disciplines from within the social and the computational sciences, as well as complex systems science (for a review on attitude change models, see \cite{Masonetal.}). One of the first works on this topic has focused on polarization, i.e. the concentration of opinions by means of interaction, as one main effect of the "social influence'' \cite{festinger50}, whereas the Social Impact Theory' \cite{Latane90} proposes a more dynamic account, in which the amount of influence depends on the distance, number, and strength (i.e., persuasiveness) of influence sources. As stated in (\cite{Castellano2007}), an important variable, poorly controlled in current studies, is structure topology. Interactions are invariably assumed as either all-to-all or based on a spatial regular location (lattice), while more realistic scenarios are ignored.  

Although very interesting, these studies fail to address the specificity of opinions, treating them as generic mental objects that change as a consequence of social influence, as it happens also to beliefs, or even goals. The question about what an opinion is and what its main features are remains unanswered, as well as their relationships with attitudes and their resistance to influencing.

\subsection{Computer science and complex systems: hyper-simplification fallacy}

Turning our attention to complex systems science, one of the most popular model applied to the aggregation of opinions is the bounded confidence model, presented in \cite{amblard01}. Much like previous studies, in this work agents exchanging information are modeled as likely to adjust their opinions only if the preceding and the received information are close enough to each other. Such an aspect is modeled by introducing a real number $\epsilon$, which stands for tolerance or uncertainty (\cite{Castellano2007}) such that an agent with opinion $x$ interacts only with agents whose opinions is in the interval $] x - \epsilon ,  x + \epsilon[$. 
This hyper-simplification helps in making this complex phenomenon more tractable using computational tools but, at the same time, reduces it to a simple exchange of values that stand for mental objects, without any kind of relationship with mental representations. An analogous attempt to model social influence has been done by Axelrod (1997), who focused on the spreading of given cultural features through communication. Again, agents do not have internal representations of what they transmit, and final results are mainly due to initial topology and to the distribution of traits, without a real exchange among agents. 

The model we present in this paper extends the bounded confidence model by providing a cognitively plausible definition of opinion as mental representations and identifying their constitutive elements and their relationships. 

We claim that opinions are highly dynamical representations resulting from the interplay of different mental objects and affected by the mental states of other individuals in the same network. 
Aim of this work is to provide an interdisciplinary account to describe how social influence leads to opinion formation, evolution and change. 
Moving from a characterization of opinions as mental representations with specific features, we will try to model how opinions are generated within the agents' minds (micro-level) and how they spread within a network of agents (macro-level). When explaining the emergence of macro-social phenomena we need to know what happens at the micro-level, i.e. what drives human actions and decisions in order to understand how individuals' representations and behaviors can give rise to socially complex phenomena and how those affect agents' actions. Without explaining how opinions are formed and manipulated within the individuals' minds, it is very difficult to account for the way in which they change as an effect of social influence. Our aim is to understand whether and how heterogeneous agents, endowed with different beliefs and goals, may come to share a given viewpoint and what consequences this sharing has on agents' behaviors. We are interested in providing answers, at least partially, to the following questions: What is an opinion? What mechanisms lead people to change their opinions? How can individuals resist to changes? What are the mechanisms of influence acting within and between individual minds? How does social impact affect agents' elaboration of new or contrasting information?

\section{A Cognitive Theory of Opinions}
This work aims at outlining a non-reductionist cognitive model of opinions and their dynamics. 
Differently from the models reviewed above, we first provide a definition of opinions as mental representations presenting specific features that make their revision and updating more or less easy and enduring. 
Moreover, grounding opinions in the minds allow us to take into account not only direct processes of revision triggered by the comparison with others' different opinions, i.e. social influence, but also revisions based upon changing in other mental representations supporting that opinion. 

The computational model introduced in this paper is intended to provide a preliminary unifying framework to define opinions and to characterize their dynamics in an easy but non-reductionist approach. Opinions in several models of opinion dynamics are considered to change according to social influence, we try to outline what is social influence and the way the social network structure affects the agents' opinions.

\subsection{Facts and evaluations: two kinds of opinions}

In everyday language the word “opinion” is often confronted with “fact”, stressing the difference between something objective because it happened and there are proofs of it, like in the latter case, and something that does not have any reference in the external reality. This distinction is important, because it points to a prominent feature of opinions, i.e. their being regarded as uncertain and not grounded in any external proof. Opinions can be debated, compared, discussed, argumented, but they can not be proven to be true, contrary to what happens with facts. However, individuals continuously resort to their opinions as less stable but more versatile mental objects whose relevance is not reduced because of their being uncertain. This feature is specific of opinions and it also explains why opinions are more prone to change and revision, especially when confronted with others' opinions. Moreover, identifying this and other traits as specific, allows us to place opinions among other kinds of mental representations, describing the kinds of relationships opinions have with epistemic and motivational mental objects.

Opinions can be described as configurations of an individual's beliefs, values and feelings that can be conditionally activated. Conditional activation points to the flexible and dynamic nature of these representations that are not grounded in certainty and that usually come out from the merging and elaboration of other representations and attitudes. Opinions are not only conditional, but also compositional.
This means that, for instance, starting from my feeling of aversion toward mathematics and as a consequence of having met a rude friend of friends who happened to teach math at school, when asked about my opinion on the time kids should spend in studying mathematics, I can form or, better, activate an opinion according to which the less time they spend the better it is. 

Opinions stem from the conditional activation of different kinds of mental representations, that can have a propositional content or, as in the case of attitudes and feelings, they can be more evaluative.
However, there is a specific feature that distinguishes an opinion from other kinds of mental objects. An opinion is an epistemic representation in which the truth-value is deemed to be uncertain. Opinions refer to objects of the external world that can not be told to be either true or false. This impossibility (or irrelevance) to say whether the content of a representation is true or false, but only if it makes sense according to what someone believes and knows is what makes a mental representation an opinion. This essential feature accounts for the fact that opinions can be easily influenced not only by social influence, i.e. an external force, but that they can also be easily revised according to the change in one's own mental representations.

This basic feature can be paired with the presence of an attitude, i.e. an evaluative component that specifies whether the individual likes or dislikes the topic. In general, attitudes are present when the topic is somehow involving for the subject, so he is positively or negatively inclined toward it.

When this is not the case, we have "factual opinions", like in the following example. If someone is required to say when Mozart died, he can know the correct answer or not, but this is not a moot point. On the contrary, the causes of Mozart's death are debatable because without knowing where he was buried it is impossible to analyze the bones and to ascertain what killed him. This means that we know that Mozart died in 1791 but there are contrasting opinions about the causes of his death,  and, even if there exist one true opinion, none can tell which is the truth.
On the other hand, when opinions involve also evaluative components or facts, the opinions result from the activation of a pattern of related representations like beliefs, knowledge, other opinions, but also goals. This view allows us to describe opinions as non-static patterns of relationships in which different representations are linked through a variety of different linkages. This work is meant to address the origin and changing of opinions thanks to these inter-relationships.

\subsection{A tripartite model of opinion: truth-value, confidence and sharedness}

An opinion is characterized by the three following features. First, the truth value can not be verified (or it is not relevant). In general, opinions are representations whose truth value can not be assessed through direct experience. The topic of the opinion can not be experienced and then it is impossible to say whether a given object is true or false. If I ask someone about his opinion on the military intervention in Afghanistan, he can not tell me that his opinion, whether positive or negative, is true, because it is not possible to test an alternative state of the world in which the intervention has not taken place and then asses which state was the best. Nonetheless, he can tell me that he has a strong opinion or that he is very confident in it because he has many supporting beliefs (e.g. Talibans' regime had to be fighted, civilians needed the intervention, the world is a safer place after the intervention, etc) and even some goals (for instance, feeling safer) related with that opinion.  We can have strong or weak opinions, but our confidence does not depend on the fact that something is known to be true, given the impossibility to assess its truth-value. In other cases, assessing the truth-value is not relevant, because the attitude and the supporting mental representations are stronger enough to support the opinion, without caring for its being true or false. Going back to the example about the time spent in studying math, I can build upon my negative experience at school, supporting it with my negative attitude and recalling my experience with the unfriendly friend of my friends who happens to be a math teacher, to build up my negative opinion. Furthermore, notwithstanding the existence of statistics or experts that can support or confute my opinion, I do not care about them, because they are not relevant to me. A creationist's opinion about Darwin and the theory of natural selection is not affected by the proofs of its validity, because he does not care for those proofs and focus his attention on other kinds of knowledge (like that coming from the Bible, for instance).

The second feature is the degree of confidence which is a subjective measure of the strength of belief and it expresses the exent to which one's opinion is resistant to change. This is to say that the lack of an assessable truth value is totally independent from the confidence one has in his opinions.The degree of confidence depends on the number of supporting representations, and the higher this number the stronger an opinion will be. Castelfranchi, Poggi \cite{CastelfranchiPoggi} made a distinction between confidence coming from the source and confidence coming from the degree of compatibility that a given belief has with pre-existing beliefs. It is interesting to notice that representations do not need to be about the same topic or to belong to the same set to form a coherent network. If we take the Afghanistan example, we can easily imagine that a negative opinion about the military intervention could be supported by a general belief about the right of other countries to intervene in internal disputes or by negative evaluations about the US foreign policy, or even by knowledge about the roles played by URSS and US in Afghanistan during the Cold War. These beliefs are not exclusively related to the target opinion and they can have stronger or weaker connections with other opinions. The stronger the confidence in these beliefs and the higher their number, the stronger will be the confidence in that opinion. 

The degree of confidence can also vary in accordance with the configuration activated by a certain opinion. Since opinions are dynamic configurations emerging from the conditional activation of other representations, the path followed to link different beliefs, goals, data and memories can result in opinions that have the same content but different degrees of confidence. I can be against the military intervention in Afghanistan because I feel empathic with the civilians, thus focusing on the attitudinal and evaluative aspects, or because I have strong beliefs about the US foreign policy. In this latter case, my opinion is supported by facts and follows a specific argumentative line, and it could lead me to be more confident.

Finally, the sharing of an opinion, i.e. the extent to which a given opinion is considered shared, is another crucial feature. The sharing may heavily affect the degree of confidence, making people feel more confident because many other individuals have the same opinion. The sharing is the outcome of a process of social influence, through which agents' opinion are circulated within the social space and they can become more or less shared. This dimension is crucial, but it is also true that it carachterizes other social beliefs, like reputation.

It is worth noticing that there are other kinds of beliefs that are really close to opinions but, at a closer investigation, there are some important differences. Reputation can be one of these, because it is shared and it is also carachterized by a varying degree of confidence. But, unlikely opinions, reputation has a truth value because it refers to someone's behaviors or actions that were actually exhibited (or that were reported as such, but we do not want to address here the issue of lying) and reported to other people. Reality matters in reputation, whereas it is much less relevant in opinions, as witnessed also by the fact that reputation does not have to be convincing (i.e. supported by some reasoning or arguments), whereas opinions need. 

\section{Toward a Formal Definition}

\subsection{Preliminaries}

\subsubsection{Time Varying Graphs}
As mentioned in previous section the temporal aspects of our opinion model is based upon Time-Varying Graphs (TVG) formalism, an algorithmic framework \cite{CFQS2010} designed to deal with the temporal dimension of networked data and to express their dynamics from an {\em interaction-centric} point of view \cite{ACFQS2010a}.

Consider a set of entities $V$ (or {\em nodes}), a set of relations $E$ between these entities ({\em edges}), and an alphabet $L$ accounting for any property such that a relation could have ({\em label}); that is, $E \subseteq V \times V \times L$.  $L$ can contain multi-valued elements. 

The relations (interactions) among entities are assumed to take place over a time dimension (continuos or discrete) $\T$ the {\em lifetime} of the system which is generally a subset of $\mathbb{N}$ (discrete-time systems) or $\mathbb{R}$ (continuous-time systems).  The dynamics of the system can subsequently be described by a time-varying graph, or TVG, \TVG, where
\begin{itemize}
\item $\rho: E \times \T \rightarrow \{0,1\}$, called {\em presence function}, indicates whether a given edge or node is available at a given time.
\item $\zeta: E \times \T \rightarrow \mathbb{T}$, called {\em latency function}, indicates the time it takes to cross a given edge if starting at a given date (the latency of an edge could vary in time).
\end{itemize}

\subsubsection{The underlying graph}
\label{sec:underlying-graph}
Given a TVG \TVG, the graph $G=(V,E)$ is called {\em underlying} graph of $\G$. This static graph should be seen as a sort of {\em footprint} of $\G$, which flattens the time dimension and indicates only the pairs of nodes that have relations at some time in a given time interval $\T$. In most studies and applications, $G$ is assumed to be connected; in general, this is not necessarily the case. Note that the connectivity  of $G=(V,E)$ does not imply that $\G$ is connected at a given time instant; in fact, $\G$ could be disconnected at all times. The lack of relationship, with regards to connectivity, between $\G$ and its footprint $G$ is even stronger: the fact that  $G=(V,E)$ is connected does not even imply that $\G$ is ``connected over time''.

\subsubsection{Edge-centric evolution}
From an edge point of view (relationships within epistemic representations), the evolution derives from variations of the availability. TVG defines the {\em available dates} of an edge $e$, noted $\I(e)$, as the union of all dates at which the edge is available, that is, $\I(e)= \{t \in \T : \rho(e,t)=1\}$. 
Given a multi-interval of availability $\I(e)=\{[t_1,t_2)\cup[t_3,t_4)...\}$, the sequence of dates $t_1,t_3,...$ is called {\em appearance dates} of $e$, noted $\AD(e)$, and the sequence of dates $t_2, t_4,...$ is called {\em disappearance dates} of $e$, noted $\DD(e)$. Finally, the sequence $t_1, t_2, t_3,...$ is called {\em characteristic dates} of $e$, noted $\ST(e)$.

\subsubsection{Graph-centric evolution}

From a global standpoint, the evolution of the system can be derived by a sequence of (static) graphs $\SG=G_{1}, G_{2}..$ where every $G_i$ corresponds to a static {\em snapshot} of $\G$ such that $e\in E_{G_i} \iff \rho_{[t_i,t_i+1)}(e)=1$, with two possible meanings for the $t_i$s: either the sequence of $t_i$s is a discretization of time (for example $t_i=i$); or it corresponds to the set of particular dates when topological events occur in the graph, in which case this sequence is equal to $sort(\cup\{\ST(e): e \in E\})$. In the latter case, the sequence is called  {\em characteristic dates} of $\G$, and noted $\ST(\G)$.

\subsection{Modeling Epistemic Representations}

An {\em opinion} is an epistemic representation of a state of the world with respect to a given object $p$. It is defined on a three dimensional space defined by: a) the {\em objective truth value} $T_o$, a {\em subjective truth value}, namely $T_s$ and a {\em degree of confidence} $d_{c}$ with respect to the object $p$.

More formally we can state that:

\begin{definition}
an epistemic representation of a state of the world $m \in M$ is a quadruplet ${p,T_o, T_s, d_{c}}$ defined by
a preposition $p$ related to a given object $O$, and two variable $T_o$ and $T_s$ defined on $\mathbb{R} $. The $d_{c} \in \mathbb{R}$ respectively quantifying the ``real`` truth value of an information, namely the objective truth value, the perceived truth values, and the degree of confidence, with respect to the preposition $p$.  
\end{definition}

By varying the dimensions of the domain of $T_o$ and $T_s$, we can define a taxonomy of the epistemic representation of the world that can be summarised as follows:

\begin{definition}
An epistemic representation $m_k = \{p,T_o,T_s,d_{c}\}$ is {\em knowledge} when $T_o = T_s$.
\end{definition}

\begin{definition}
An epistemic representation $m_b = \{p,T_o,T_s,d_{c}\}$ is a {\em belief} when $0 < T_o < 1 
\wedge 0 \leq T_s \leq 1$ .
\end{definition}

\begin{definition}
An epistemic representation $m_o = \{p,T_o,T_s,d_{c}\}$ is an \textbf{opinion} when $0 \leq T_o < 1 \wedge 0 \leq T_s \leq 1$.
\end{definition}

\subsection{Opinions and Individuals}

We can define an epistemic representation graph as a network of epistemic representation immerged in a dynamic network in a given time interval and the links state the correlation among them.
Let us consider  a  set  $V$ of mental representation (or nodes), interacting with one another over time.
Each {\em relation} among the mental representation can be formalized by a quadruplet $c=\{u,v,t_1,t_2\}$, where $u$ and $v$ are the involved mental representations (either beliefs, or knowledge or an opinion), $t_1$ is the time at which the correlation occurs, and $t_2$ the time at which the relation terminates.  A given pair of nodes can naturally be subject to several such interactions over time (and for generality, we allow these interactions to overlap). Given a time interval $\T=[t_a, t_b)\subseteq {\cal T}$ (where $t_a$ and $t_b$ may be either two dates, or one date and one infinity, or both infinities), the set $C(\T)$ (or simply  $C$) of all interactions  occurring during that time interval defines a set of intermittently-available edges $E(\T) \subseteq V\times V$, such that:

\begin{equation}
\begin{split}
  \forall u,v \in V, (u,v) \in E(\T) \\
\iff \exists t' \in [t_a,t_b), (u,v,t_1,t_2)\in C(\T) \ : \ t_1\le t'< t_2
\end{split}
\end{equation}

 \noindent that is, an edge $(u,v)$ exists iff at least one interaction between $u$ and $v$ occurs, or terminates, between $t_a$  and $t_b$. The intermittent availability of an edge $e=(u,v)\in E(\T)$ is described by the {\em presence function} $\rho: E(\T)\times \T \rightarrow \{0,1\}$ such that
  $ \forall t \in \T, e\in E(\T)$:

\begin{equation}
 \rho(e, t)=1 \iff    \exists  (u,v,t_1,t_2)\in C :  t_1\le t<t_2
\end{equation}

The triplet $\G=(V,E,\rho)$ is called an {\em epistemic representation graph}, and
the temporal domain  $\T=[t_a, t_b)$ of the function $\rho$, is  the {\em lifetime} of $\G$. 
We denote by $\G_{[t,t')}$  
the {\em mental representation subgraph} of $\G$ covering the period $[t_a,t_b) \cap [t,t')$

Hence, a sequence of couples $\J=\{(e_1,t_1), (e_2, t_2),...\}$, with $e_i \in E$ and $t_i \in \T$ for all $i$, is called a {\em journey} in $\G$ iff $\{e_1, e_2,...\}$ is a walk in $G$ and for all $i$, $\rho(e_i, t_i)=1$ and $t_{i+1} \ge t_{i}$. Journeys can be thought of as {\em paths over time} from a source node to a destination node (if the journey is finite).

Let us denote by $\J^*_\G$ the set of all possible journeys in an epistemic representation system $\G$. We will say that $\G$ {\em admits} a journey from a node $u$ to a node $v$, and note $\exists \J_{(u,v)} \in \J^*_\G$, if there exists at least one possible journey from $u$ to $v$ in $\G$.

\subsection{Opinion Dynamics and Society}

One of the most famous formalisms aimed at describing the process of persuasion is the ``Bounded Confidence Model'' (BCM) where agents exchanging information are modeled as likely to adjust their opinions only if the preceding and the received information are close enough to each other. Such an aspect is modeled by introducing a real number $\epsilon$ , which stands for tolerance or uncertainty  such that an agent with opinion $x$ interacts only with agents whose opinions is in the interval ]x − $\epsilon$ , x + $\epsilon$ [. 
Neverthless the wide, massive and cross-disciplinary use of the BCM (\cite{Lorenz07,Hu09}) ranging from ``viral marketing'' to to the Italians' opinions distortion played by controlled mass media (\cite{quattrociocchi2010e,brunetti2010,brunetti2010a,Hu09}). Such a model does not provide an explanation of the phenomena yielding to the tolerance value, it is just assumed as a static value.

In this work we will outline which are the factors affecting the acceptance or the refuse of one another opinion. 
In particular, how can we formalize comparison of two or more opinions? Recalling that a mental representation is a preposition with the truth value defined by two variable $T_o,T_s \in \mathbb{R} $ and $d_{c} \in \mathbb{R}$ respectively quantifying the ``real'' and the perceived truth value and the degree of confidence with respect to a given object or proposition.
And considering that such mental representations are modeled as set of time connected entities of the form $\G=(V,E,\rho)$ we can now provide some definitions aimed at describing the process of persuasion.

Assuming that an epistemic representation system, which is by nature adaptive, when facing with external events, reacts to the stimulus by activating only a subset of its components. For instance, consider the example where an agent $x$ is questioned by an agent $y$ about his opinion on a given target.

What does happen in the $x$'s mental representation system? How can we quantify $x$'s attitudes to change or not is opinions regarding a given matter of fact?

According to our model the epistemic representation system of $x$, as reaction to the external stimulus posed by the $y$'s question, will perform $journey$ within the elements that in its mind are related with the target of the question and on this base will be able to compare its opinion with the one owned by $y$.

\begin{definition}
{\em (relational-)connected component induced by an external event} in $\G_{x}$ is defined as a set of nodes $V'\subseteq V$ such that $\forall u,v \in V', \exists \J_{(u,v)} \in \J^*_\G$. Then $\G$ is said connected if it is itself a connected component ($V'=V$). 
\end{definition}

Since all nodes in $V'$ are defined by an objective truth value $T$ and a degree of confidence (perceived truth value) $d_g$ it is obvious that the resistence to an opinion to change is denoted by these values in all the nodes in $V'$.

\section{Conclusions}
\label{sec:5}

In this preliminary work we tried to sketch a cognitively grounded dynamic model of opinions, in which we defined these mental representations as carachterized by the presence of three specific features. Differently than psychological theories of opinions that usually provide rich definitions that are too complex to be reduced to measurable variables, we isolated three main constitutive elements that characterize this kind of mental representations. On the other hand, we tried to overcome the reductionist approach of opinion dynamic models, in which the richness of human cognitive processes is substituted by easy-to-compute factors poorly related to actual human behaviors. For this reason, we proposed to apply time-varying-graph to develop a formal model able to account for the way in which opinions are generated and change as a function of the presence and opinions of other agents in the network.

We are perfectly aware of the complexity of this issue and this work represents a preliminary attempt to merge the cognitive complexity of opinions with a rigorous formal approach, but there are many problems that we need to address. First, the cognitive model should be refined and specific hypotheses about opinion revision and diffusion should be put forward. Moreover, the robustness of the formal model will be tested and such a model will be implemented in cognitive multi-agent system in order to explore the parameter space upon which our model has been defined. Our ultimate aim is to build up a simulation environment in which agents endowed with heterogeneous representations of the external world interact and this leads to the creation of new opinions, the disappearing of some of the previous ones and, in general, to different distributions of representations in the population.

\section{Acknowledgements}
This work was supported by the European Community under the FP6
programme (eRep project CIT5-028575). A particular thanks to Ilvo Diamanti, Federica
Mattei, Mario Paolucci, Federico Cecconi, Stefano Picascia, Geronimo Stilton and the Hypnotoad.
In addition we are grateful to the biggest Italian anomaly and the Italian media for the inspirations and insights.

\bibliographystyle{plain}
\bibliography{biblio}

\end{document}